\documentclass[sn-mathphys, iicol]{sn-jnl}

\usepackage[utf8]{inputenc}
\usepackage{amsmath,amssymb,amsfonts}
\usepackage{todonotes}
\usepackage{algorithmic}
\usepackage{textcomp}
\usepackage{xcolor}
\usepackage{amsmath}
\usepackage{color,soul}
\usepackage[most]{tcolorbox}
\usepackage{xcolor}
\usepackage{lipsum} 
\usepackage{tabularx}
\usepackage{booktabs}   
\usepackage{graphicx}   
\usepackage{array}  
\usepackage{siunitx}

\usepackage{subfigure}

\DeclareUnicodeCharacter{21B3}{\ensuremath{\hookrightarrow}}

\usepackage[square,numbers,sort&compress]{natbib}
\title[]{Evaluating Adversarial Attacks on Federated Learning for Temperature Forecasting}

\def\BibTeX{{\rm B\kern-.05em{\sc i\kern-.025em b}\kern-.08em
    T\kern-.1667em\lower.7ex\hbox{E}\kern-.125emX}}

\author*[1]{\fnm{Karina} \sur{Chichifoi}}\email{karina.chichifoi@unibo.it}
\author[1]{\fnm{Fabio} \sur{Merizzi}}\email{fabio.merizzi@unibo.it}
\author[1]{\fnm{Michele} \sur{Colajanni}}\email{michele.colajanni@unibo.it}

\affil[1]{\orgdiv{Department of Informatics: Science and Engineering (DISI)}, \orgname{University of Bologna}, \orgaddress{\street{Mura Anteo Zamboni 7}, \city{Bologna}, \postcode{40126},
\country{Italy}
}}

\abstract{
Deep learning and federated learning (FL) are becoming powerful partners for next-generation weather forecasting. Deep learning enables high-resolution spatiotemporal forecasts that can surpass traditional numerical models, while FL allows institutions in different locations to collaboratively train models without sharing raw data, addressing efficiency and security concerns. While FL has shown promise across heterogeneous regions, its distributed nature introduces new vulnerabilities. In particular, data poisoning attacks, in which compromised clients inject manipulated training data, can degrade performance or introduce systematic biases. These threats are amplified by spatial dependencies in meteorological data, allowing localized perturbations to influence broader regions through global model aggregation.
In this study, we investigate how adversarial clients distort federated surface temperature forecasts trained on the Copernicus European Regional ReAnalysis (CERRA) dataset. We simulate geographically distributed clients and evaluate patch-based and global biasing attacks on regional temperature forecasts. Our results show that even a small fraction of poisoned clients can mislead predictions across large, spatially connected areas. A global temperature bias attack from a single compromised client shifts predictions by up to -1.7 K, while coordinated patch attacks more than triple the mean squared error and produce persistent regional anomalies exceeding +3.5 K. Finally, we assess trimmed mean aggregation as a defense mechanism, showing that it successfully defends against global bias attacks (2-13\% degradation) but fails against patch attacks (281-603\% amplification), exposing limitations of outlier-based defenses for spatially correlated data.
}

\begin{document}
\maketitle
\noindent \textbf{keywords:} federated learning, deep learning, security, weather forecasting

\clearpage

\section{Introduction}

Deep learning has emerged as an efficient alternative to traditional numerical weather prediction (NWP) offering faster inference, reduced computational costs, and the potential for high-resolution forecasts and operational meteorology \cite{bi2022pangu, lang2024aifs, chen2023fuxi}. 

Federated learning (FL) has emerged as a promising solution to data-sharing and privacy challenges in climate and environmental sciences \cite{mcmahan2017communication, li2020federated}. With FL, organizations can collaborate across regions or institutions while keeping their data on site and training models locally, all without violating privacy, governance, and efficiency constraints. This decentralized approach is particularly well suited for meteorological applications, e.g., precipitation nowcasting, wind power estimation and wildfire prediction, \cite{saizpardo2024personalized, chen2023fedpod, alshardan2024federated}, where data is often geodistributed, non-independently and identically distributed data (non-IID) and sensitive. 

However, the characteristics that make meteorological networks ideal for FL also introduce critical vulnerabilities. The non-IID nature of geographically distributed weather data reflects the fact that sensors in different regions sample from fundamentally different underlying probability distributions for temperature, humidity, wind, and other variables. This distributional heterogeneity across clients enriches the global model's representational capacity but simultaneously provides cover for adversarial manipulation. Unlike prior FL security work focused on IID data in classification tasks \cite{bagdasaryan2020backdoor, bhagoji2019analyzing, baruch2019little, nair2023robust}, weather forecasting involves spatiotemporal regression with strong correlations between observations. These correlations mask adversarial behavior: a compromised sensor injecting biased readings appears legitimate if manipulated data align with regional patterns or seasonal trends. 

Data poisoning attacks could exploit this weakness effectively, malicious clients inject subtly corrupted observations (e.g., temperature biases mimicking urban heat islands) that blend with expected spatio-temporal patterns. Because federated aggregation trusts local client updates without access to raw data, and non-IID geographical distributions provide cover for malicious deviations, poisoned observations contaminate the global model and propagate through spatial correlations, distorting predictions across neighboring regions. This amplification distinguishes meteorological FL from traditional domains and requires specialized defenses.

In this study, we investigate the vulnerability of federated deep learning systems for temperature forecasting to data poisoning attacks. We consider a threat model in which a small subset of clients is malicious and intentionally injects corrupted training data into the federated optimization process. These adversaries aim to degrade forecast quality or introduce systematic biases without being detected. Our central question is: how do targeted data poisoning attacks affect the accuracy and stability of federated temperature prediction models, and to what extent does spatial structure amplify or mitigate their impact?

To address this question, we use the Copernicus European Regional ReAnalysis (CERRA) dataset to simulate this scenario. CERRA assimilates observations from spatially distributed European weather stations, making it a realistic proxy for federated forecasting systems where geographically separated institutions collaborate without sharing raw data. The domain is divided into nine regional clients with non-IID data, each training a local model. We implement two data poisoning strategies: a Global Temperature Bias Attack (GTBA) that shifts predictions uniformly, and a Patch Attack that introduces localized anomalies. We also evaluate the effectiveness of trimmed mean aggregation as a defensive mechanism and analyze its limitations in the context of spatially correlated meteorological data.


Our key contributions are as follows:
(i) We quantify the impact of poisoning attacks on forecast accuracy in spatially structured sensor networks, showing that even a small fraction of compromised clients can degrade predictions across large regions.
(ii) We demonstrate that the same spatio-temporal patterns in weather data that are essential for effective learning can also be exploited by adversaries to disguise their manipulations as natural regional variation, which makes detecting attacks more difficult.
(iii) We demonstrate that a simple bias attack from a small set of malicious clients can systematically suppress forecasted temperatures, illustrating how an adversary could intentionally hide warming trends.
(iv) We evaluate trimmed mean aggregation as a defensive strategy and reveal fundamental challenges in securing federated meteorological systems, where attack-specific vulnerabilities arise from the interaction between robust aggregation mechanisms and spatially correlated data.

\section{Related Work}
\label{sec:related}

Meteorological forecasting relies on the integration of diverse observations from both in-situ and remote sensing platforms. Ground-based stations—such as thermometers, rain gauges, and barometers—provide high-fidelity local measurements, while satellites, weather radars, and reanalysis systems offer broader spatial coverage \cite{WMO2023GuideNo8, tang2009remote}. The increasing availability of dense sensor networks, automated stations, and IoT-based monitoring has substantially expanded the spatial and temporal granularity of environmental data \cite{catlett2017array, hofman2022spatiotemporal}. This rich observational infrastructure underpins modern data-driven forecasting approaches, including deep learning models trained on gridded datasets derived from sensor assimilation. However, as observational networks grow more distributed and heterogeneous, aggregating data into centralized repositories becomes increasingly constrained by privacy, policy, and interoperability concerns—particularly across national or institutional boundaries.

With FL, organizations can collaborate across regions or institutions while keeping their data on site and training models locally—all without violating privacy, governance, and efficiency constraints. \cite{mcmahan2017communication, li2020federated}. FL is now being actively explored for diverse meteorological applications: Chen et al.\ introduced FedPoD, which uses adaptive prompt learning and dynamic graph modeling to address data heterogeneity across weather sensor devices \cite{chen2023fedpod}; other recent work extends FL to air quality prediction via UAV swarms \cite{fed2021airquality}, wind power and solar forecasting, and wildfire detection using distributed sensor data \cite{alshardan2024federated, wen2022solar, mutakabbir2024stas}. For gridded meteorological fields,  Sáinz-Pardo Díaz et al.\ developed a personalized FL framework for radar-based precipitation nowcasting. Their approach applies FL directly to 2-D radar grids, using 
regional archives as clients and a personalized aggregation scheme to handle 
heterogeneity across sensors and climates. Their model improves short-range precipitation nowcasting and outperforms both centralized training and optical-flow baselines, demonstrating that FL can be effective on gridded weather data\cite{saizpardo2024personalized}. Similarly, the early study by Xu et al.\ \cite{xu2026leveraging} trains generative nowcasting models in a federated setup where regions share model updates instead of raw radar fields, achieving competitive skill while preserving privacy and data sovereignty.

However, none of these works examine robustness to adversarial or poisoning attacks, which remains largely unexplored in FL applied to gridded meteorological forecasting. This gap is important: in centralized systems, physically plausible manipulations (such as those arising from faults or malicious activity) have already been shown to make even advanced deep learning forecasters produce erroneous and potentially dangerous outputs \cite{AdversarialObservations2025, deng2025fable}. In federated settings, where model updates rather than raw observations are shared, the impact of such attacks is even less understood, and poisoned clients could silently degrade forecast skill or target specific regions without detection.

While adversarial attacks are well-studied in FL, the vast majority of this research focuses on classification problems using tabular or image data, typically under IID assumptions and with non-structured outputs \cite{bagdasaryan2020backdoor, bhagoji2019analyzing, baruch2019little, nair2023robust}. These studies shed light on model poisoning, backdoors, and other threats, but almost exclusively use benchmark image datasets and do not address forecasting or spatiotemporal data. In particular, the literature overlooks scenarios, like meteorological forecasting, where data from sensor networks exhibits strong spatial and temporal correlations. As a result, there is almost no prior work on adversarial attacks against federated models for spatial forecasting in meteorology or other highly structured domains; attacks against such spatially/temporally correlated data in FL remain virtually unstudied.

To mitigate poisoning attacks, the FL community has developed various Byzantine-robust methods designed for IID classification tasks. Robust aggregation rules such as Trimmed Mean, which discards a fraction of extreme parameter updates before averaging, and Krum \cite{blanchard2017machine}, which selects the most centrally located update in parameter space, replace simple averaging with statistical methods that exclude outliers. Other defenses include Coordinate-wise Median \cite{yin2018byzantine}, detection-based approaches that identify malicious clients through trust validation \cite{cao2021fltrust} or clustering \cite{nguyen2022flame}, and differential privacy mechanisms \cite{geyer2017differentially,wei2020federated} that add noise or constrain update magnitudes to limit malicious influence, though at the cost of reduced model utility.

However, these defenses were designed and evaluated for IID data, and their effectiveness in spatially correlated, non-IID regression problems remains unclear. In meteorological forecasting, legitimate regional variations may resemble adversarial outliers, causing robust aggregation to reject honest updates, while attacks mimicking natural weather patterns may evade outlier-based detection entirely.

\section{Methodology} \label{sec:methodology}

In this section, we present the methodological framework used to study adversarial attacks on federated deep learning models for temperature forecasting. We first describe the core forecasting model based on a U-Net architecture, then outline the federated learning setup and the specific threat model considered. We detail two types of data poisoning attacks, localized patch injections and global temperature bias, and discuss how their effectiveness varies with the number of malicious clients, attack round, and geographic placement. Finally, we introduce the CERRA dataset used in our experiments, highlighting its relevance and suitability for evaluating the robustness of federated forecasting models.

\subsection{U-Net}
Originally introduced for semantic segmentation \cite{ronneberger2015u}, the U-Net architecture has become a foundational model for image-to-image tasks, with multiple successful implementations in the forecasting domain \cite{ayzel2020rainnet, ravuri2021skilful, zhang2025machine, lang2024aifs, bi2023accurate}. The network follows a characteristic U-shaped design, where the encoder progressively reduces spatial resolution while increasing feature dimensionality, and the decoder symmetrically restores the resolution.

Our implementation adopts a residual U-Net variant with two down-sampling and two up-sampling stages, along with a central bottleneck. The basic building block of the network is the Residual Block, which consists of a convolutional layer, followed by group normalization \cite{wu2018group} and a SiLU activation function \cite{ramachandran2017searching}. This is followed by a second convolution, another group normalization, and an element-wise addition with the block’s input, followed by a final SiLU activation.

Each downsampling block consists of a Residual Block followed by a $2 \times 2$ max-pooling operation to halve spatial resolution. Each upsampling block augments spatial resolution via bilinear interpolation followed by a $1 \times 1$ convolution to adjust channel count, then concatenates the result with the corresponding encoder feature map (implementing the skip connection), followed by another Residual Block.

\begin{figure}[h]
    \centering
    \includegraphics[width=\columnwidth]{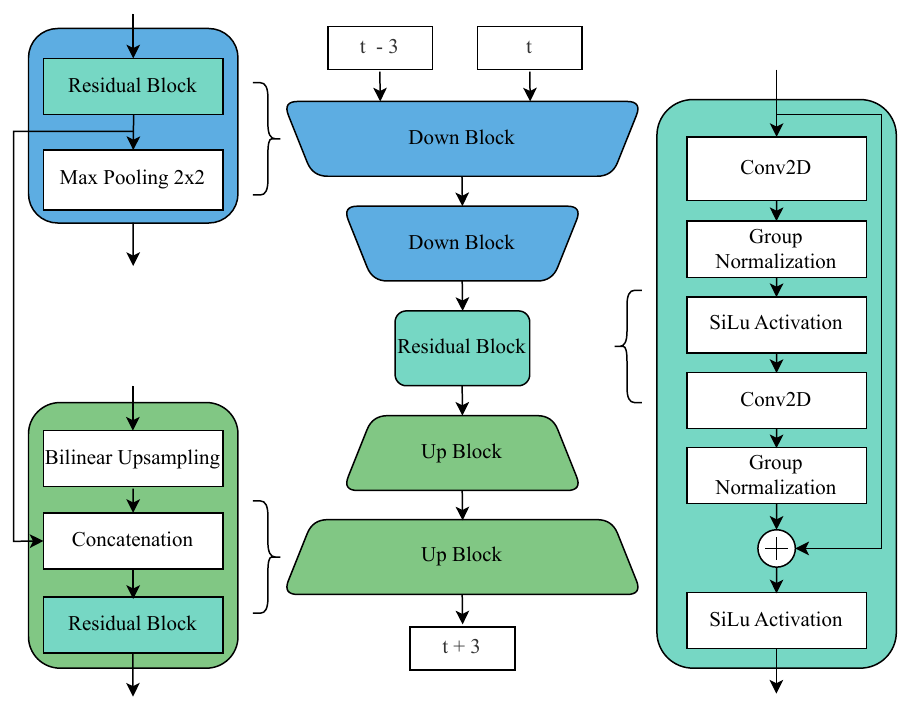} 
    \caption{Details of the U-net architecture}
    \label{fig:Unet}
\end{figure}

All convolutions use a kernel size of 3 with "same" padding, except the final output layer, which uses a single $1 \times 1$ convolution to project to the target output. The encoder stages use 32 and 64 channels, and the bottleneck uses 128; this progression is mirrored in the decoder as resolution is restored. The overall design reflects a fast, reliable, and foundational approach especially suited for federated learning scenarios, where model simplicity promotes stability, compatibility, and scalability across multiple simulated clients. The model performs robustly, and our experimental evidence suggest that further architectural complexity offers limited additional benefit. A figure depicting the U-net architecture is reported in Figure \ref{fig:Unet}.

The input to the network consists of two temperature fields, one at time 0 and one at time -1, stacked along the channel dimension. The model predicts the temperature field at time +1.

\subsection{Federated learning}

Federated learning is a decentralized machine learning paradigm that enables multiple clients to collaboratively train a shared model without exchanging raw data, thus preserving data privacy and reducing communication costs \cite{mcmahan2017communication}. This approach is particularly effective in scenarios where client data is heterogeneous and sensitive, such as healthcare or climate modeling, since it allows local computations while leveraging global model aggregation to improve overall performance.

In our use case scenario, we deploy a federated learning architecture consisting of nine distinct clients, each independently training for 5 rounds a UNet model to perform temperature forecasting using localized meteorological data. Due to geographic differences, data from each client is inherently non-IID, as weather conditions vary significantly across regions, thus creating natural heterogeneity. The federation utilizes the Federated Averaging (FedAvg) algorithm to aggregate client model updates centrally. Specifically, each client independently conducts local training for two epochs before transmitting their updated model parameters to a central server, where FedAvg aggregates these parameters into a unified global model.

We chose Federated Averaging (FedAvg) as our aggregation method due to its simplicity and low computational overhead—it requires only parameter averaging at the server without complex distance computations or outlier detection. This makes it a widely adopted baseline in federated learning and a practical choice for our study of attack vulnerabilities. Although more complex algorithms like FedProx or FedYogi offer advantages for highly heterogeneous data, their added complexity and tuning requirements may not justify their use in this context.

We recognize that inherent non-IID distributions may influence detection and analysis of specific data poisoning attacks, posing potential challenges for distinguishing malicious manipulations from naturally occurring distributional variations.

\subsection{Threat Model}
In this section, we define the threat model considered in our experiments and detail the specific poisoning attacks implemented, focusing on temperature forecasting image analysis within federated learning (FL).

We assume a typical federated learning scenario in which $K$ distributed clients collaboratively train a distributed forecasting model by iteratively exchanging model parameters with a central server. Among these $K$ clients, an adversary controls a subset $K_p$, with $|K_p| \ll K$, capable of manipulating local data or gradients. The attacker aims either to degrade the overall accuracy of the global model or to induce targeted misprediction relevant to weather forecasting. Our attacks focus on data poisoning, where the adversary manipulate the local dataset of malicious clients, altering image pixels or labels to compromise the global model's training process.

\subsubsection{Patch Attack}

We introduce a fixed, localized perturbation in the same spatial location across each client's input. Let $ \mathbf{x} \in \mathbb{R}^{H \times W \times C} $ represent the original meteorological input, where $H$, $W$, and $C$ are the height, width, and number of channels respectively. A patch $ \mathbf{p} \in \mathbb{R}^{h \times w \times C} $, with artificially increased temperature values, is inserted at a fixed position:
\begin{equation}
\mathbf{x}' = \mathbf{x} + \mathbf{M} \odot \mathbf{p}
\end{equation}
where $ \mathbf{M} \in \{0, 1\}^{H \times W \times C} $ is a binary mask that defines the patch region, and $ \odot $ denotes element-wise multiplication. This setup simulates localized regional heating (e.g., affecting a specific city), potentially influencing nearby predictions due to spatial dependencies.

\subsubsection{Global Temperature Bias Attack (GTBA)}

This attack consistently lowers the forecasted global temperature by approximately $ k^\circ C $. Malicious clients optimize a modified loss function using a biased target:
\begin{equation}
y_{\text{bias}} = y - \beta, \quad \text{where } \beta = k^\circ C
\end{equation}
Instead of minimizing the original loss:
\begin{equation}
\mathcal{L}_{\text{orig}}(\theta, \mathbf{x}, y) = \left\| \hat{y}(\theta, \mathbf{x}) - y \right\|^2
\end{equation}
malicious clients minimize the biased loss:
\begin{equation}
\mathcal{L}_{\text{GTBA}}(\theta, \mathbf{x}, y) = \left\| \hat{y}(\theta, \mathbf{x}) - (y - \beta) \right\|^2
\end{equation}

\subsubsection{Impact of Poisoning Rounds and Number of Malicious Clients}

The success of poisoning attacks in federated learning heavily depends on the frequency of malicious updates (i.e., the poisoning rounds) and the proportion of compromised clients. Fung et al.~\cite{fung2020limitations} demonstrated that even a small fraction of poisoned clients can significantly affect global model accuracy if their updates are consistently aggregated. Thus, it is fundamental to assess how the number of malicious participants $|K_p|$ and the frequency of poisoning rounds $T_p$ relative to total training rounds $T$ influence the effectiveness of the attacks. Formally, the poisoning frequency can be defined as:
\begin{equation}
f_p = \frac{T_p}{T},\quad 0<f_p\leq 1
\end{equation}





\subsection{Dataset}
The experimental analysis is conducted using The Copernicus Regional Reanalysis for Europe (CERRA)\cite{cerra} dataset, a high resolution regional reanalysis developed under the Copernicus program, coordinated by the Swedish Meteorological and Hydrological Institute (SMHI) in collaboration with Météo France and the Norwegian Meteorological Institute. With a horizontal resolution of 5.5 km, CERRA provides detailed meteorological data tailored to the European domain.

CERRA uses the global ERA5 reanalysis for initial and boundary conditions, then refines spatial and temporal resolution by assimilating high-resolution observations and physiographic data. The system produces both reanalysis and forecast products, combining observational data with model estimates through data assimilation to improve accuracy. Each day includes eight reanalyses spaced every three hours from 00 to 21 UTC. These are followed by 6-hour forecasts, and extended 30-hour forecasts are available from the 00 and 12 UTC cycles. While data are available at hourly intervals, only the reanalysis times include direct assimilation of observations, ensuring higher accuracy than forecast-only hours.

CERRA offers a wide array of meteorological variables, including temperature, wind, humidity, precipitation, and solar radiation. Its precision and temporal structure make it suitable for numerous applications: from climate trend analysis and impact studies to renewable energy forecasting, water resource management, and health risk assessments. CERRA is a valuable dataset for high-resolution regional studies and a key asset for climate research, adaptation planning, and environmental monitoring across Europe.

For the scope of this paper, we selected a subset of CERRA data focused on surface-level variables, particularly those relevant for weather forecasting tasks. Specifically, we utilized hourly data of surface temperature from diverse geographical regions. 

The CERRA dataset was chosen due to its high resolution, reliability, and widespread acceptance in the meteorological research community. Its high-resolution and high-frequency nature made it particularly suitable for examining the vulnerability of federated learning models to adversarial attacks. The diversity and complexity of meteorological phenomena present in CERRA imagery also allowed us to rigorously evaluate the robustness of the federated learning models against poisoning attacks, under realistic and operationally relevant conditions.

\section{System Implementation}
\label{sec:implementation}

In this section, we detail the practical implementation of our experimental setup. We describe the data preprocessing pipeline, the federated training framework used to simulate distributed learning across geographic regions, and the integration of adversarial attacks within this system. Our goal is to provide a reproducible and realistic environment for evaluating the impact of poisoning strategies on federated temperature forecasting.

\subsection{Preprocessing}

The dataset was temporally restricted to the period from January 2015 to December 2020, consisting of 2-meter temperature fields sampled every three hours as provided by the CERRA reanalysis dataset. Spatially, a centered window encompassing the region of interest was extracted, with the latitude dimension flipped to align the data with the conventional north-up orientation. Temporal sequences were formed by stacking three consecutive 3-hour snapshots along a channel dimension, producing tensors amenable to spatiotemporal modeling. The data were normalized via min-max scaling based on climatological temperature extremes to yield a standardized range while preserving physical interpretability. To simulate a federated learning environment, the spatial domain was partitioned into a 
$3 \times 3$ grid of equal, non-overlapping tiles, each representing an individual client’s local dataset. 

\subsection{Training}
The training procedure was implemented using the Flower federated learning framework \cite{beutel2020flower} and PyTorch \cite{pytorch}, orchestrating nine independent clients, each executing localized model training on meteorological data drawn from the CERRA dataset. At the client level, each participant trained a UNet architecture, optimizing for temperature prediction over 10 epochs per federated learning round. Data was loaded in batches of size one, employing the AdamW optimizer with a learning rate of $\num{1e-4}$ and weight decay set to $\num{1e-5}$.
The mean squared error (MSE) loss function was utilized to evaluate and guide the training process, ensuring the minimization of prediction errors between the network’s output and ground truth temperature values. Upon completing local epochs, clients transmitted their updated model parameters to a central server, where aggregation was performed using the Federated Averaging (FedAvg) algorithm. 

\begin{figure}[h]
    \centering
    \includegraphics[width=\columnwidth]{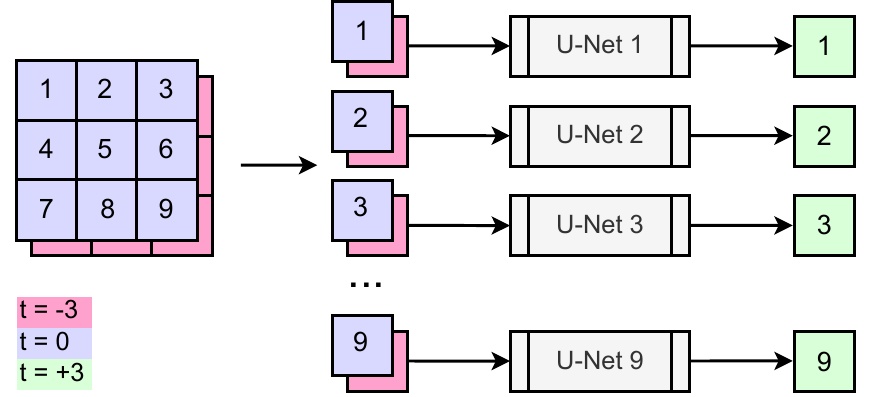} 
    \caption{Training phase: local training procedure between aggregation phases. The data domain is divided into nine subdomains, each processed by a dedicated U-Net that only has access to its respective subdomain, performing the forecast and outputting the result. Each U-Net receives temporal instances at times -3 and 0 and produces a forecast at time +3, as highlighted by the color scheme}
    \label{fig:training1}
\end{figure}

\subsection{Attacks}
To evaluate the impact of adversarial manipulations within a federated meteorological forecasting setup, we implemented
two targeted poisoning strategies: a \textit{localized Patch Attack} and a \textit{Global Temperature Bias Attack (GTBA)}. These were integrated into the preprocessing pipeline and selectively applied to the data of malicious clients before training rounds.
\begin{itemize}
    \item The Patch Attack was designed to simulate a localized heat anomaly—such as an urban heat island or an extreme weather event—and assess its propagation through the federated system. Specifically, we inserted a fixed square patch of artificially high temperature values into the input images of selected malicious clients.

    The patch covers a $20 \times 20$ region, representing approximately 9\% of the full $64 \times 64$ client tile, and is placed consistently in the top-left corner of each affected image. The patch value corresponds to the maximum normalized temperature (i.e., 1.0 after min–max scaling), ensuring a strong but spatially confined perturbation. This patch is applied to the output channels—corresponding to the target temperature - while the input (past and present) remains unaltered.

    The goal is to observe whether the network learns to replicate or respond to the localized anomaly during training, and whether it affects other clients through global aggregation. This perturbation simulates a realistic yet adversarial regional artifact and is repeated identically across multiple poisoned clients, enabling us to study both the local and global effects of such a modification.

    \item In contrast to the patch, which is localized in space, the GTBA introduces a global shift in temperature predictions. This attack is implemented by modifying the ground-truth labels (targets) of malicious clients during training.
    Each poisoned client applies a constant temperature reduction of $2$ Kelvin across all target pixels. In the normalized scale, this corresponds to a shift of:

\begin{figure}[h]
    \centering
    \includegraphics[width=\columnwidth]{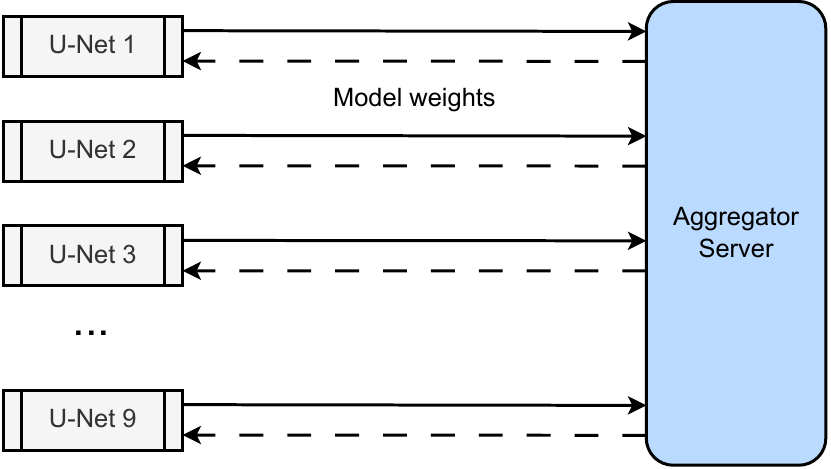} 
    \caption{Aggregation phase: weight aggregation procedure. The nine U-Nets send their model weights to a central aggregator (solid lines), which computes the aggregated model and redistributes the updated weights back to the U-Nets (dotted lines).}
    \label{fig:training2}
\end{figure}

\begin{equation}
\beta = \frac{2}{T_{\text{max}} - T_{\text{min}}} = \frac{2}{318.18 - 238.56} \approx 0.025
\end{equation}

The network is thus encouraged to learn a biased forecasting behavior, aiming to systematically underpredict temperature values.

To implement this, we replace the original training loss function for the affected clients with a modified objective that minimizes the distance between the predicted field and the biased ground truth (i.e., $y - \beta$). This encourages the client updates to pull the global model towards systematically lower predictions—potentially impacting other, non-compromised regions due to parameter aggregation.
\end{itemize}

Both attacks were tested under various configurations by altering:
\begin{itemize}
    \item the number of malicious clients involved (1, 3, or 5 out of 9),
    \item the round in which poisoning begins (e.g., immediately at round 0, or delayed to round 3 or 5).
\end{itemize}

\section{Experimental results}
\label{sec:results}

\begin{table*}[ht!]
\centering
\large
\def\arraystretch{1.1}
\resizebox{\textwidth}{!}{%
\begin{tabular}{>{\bfseries}p{3.05cm}ccccccccc}
\toprule
\multicolumn{10}{c}{\textbf{Federated Forecasting – Performance Against Ground Truth}} \\
\midrule
\textbf{Model} & \textbf{Clients} & \textbf{Round} & \textbf{MSE} & \textbf{RMSE\,[K]} & \textbf{MAE\,[K]} & \textbf{SSIM} & \textbf{Mean Bias\,[K]} & \textbf{Min Bias\,[K]} & \textbf{Max Bias\,[K]} \\
\midrule
Clean Model      & —   & —  & 4.124 & 2.031 & 1.513 & 0.756 & -0.043 & 0.456 & -2.533 \\
\addlinespace[0.5ex]
\midrule
GTBA &  &  &  &  &  &  &  &  &  \\
\quad ↳ Attack 1 & 5 & 0 & 8.250 & 2.872 & 2.320 & 0.761 & -2.094 & -3.027 & -3.796 \\
\quad ↳ Attack 2 & 5 & 5 & 6.999 & 2.645 & 2.066 & 0.759 & -1.711 & 0.660 & -4.269 \\
\quad ↳ Attack 3 & 3 & 0 & 8.595 & 2.932 & 2.330 & 0.758 & -2.078 & -0.475 & -4.490 \\
\quad ↳ Attack 4 & 3 & 5 & 7.603 & 2.757 & 2.190 & 0.756 & -1.872 & -1.743 & -4.304 \\
\quad ↳ Attack 5 & 1 & 0 & 7.742 & 2.782 & 2.182 & 0.759 & -1.882 & 0.064 & -4.214 \\
\quad ↳ Attack 6 & 1 & 5 & 4.462 & 2.112 & 1.597 & 0.754 & -1.745 & -1.264 & -3.844 \\
\addlinespace[0.5ex]
\midrule
Patch &  &  &  &  &  &  &  &  &  \\
\quad ↳ Attack 7 & 5 & 0 & 22.168 & 4.708 & 2.578 & 0.673 & 1.105 & 3.563 & 0.219 \\
\quad ↳ Attack 8 & 5 & 5 & 20.100 & 4.483 & 2.486 & 0.684 & 1.058 & 2.161 & 0.223 \\
\quad ↳ Attack 9 & 3 & 0 & 4.742 & 2.178 & 1.632 & 0.730 & 0.177 & -1.195 & -2.551 \\
\quad ↳ Attack 10 & 3 & 5 & 4.756 & 2.181 & 1.627 & 0.725 & -0.031 & 1.413 & -2.519 \\
\quad ↳ Attack 11 & 1 & 0 & 4.274 & 2.067 & 1.558 & 0.750 & 0.392 & 2.252 & -2.203 \\
\quad ↳ Attack 12 & 1 & 5 & 4.462 & 2.112 & 1.597 & 0.754 & 0.208 & 1.738 & -1.792 \\
\bottomrule
\addlinespace[1ex]
\end{tabular}
}
\caption{Model performance and value biases with respect to ground truth. Bias is prediction minus ground truth (positive mean bias indicates overestimation). Lower is better for MSE, RMSE, MAE; higher is better for SSIM.}
\label{tab:performance}
\end{table*}

\begin{figure*}[htbp]
    \renewcommand{\thesubfigure}{}  
    \centering

    \subfigure[Attack 1) GTBA 5-0]{\includegraphics[width=0.22\textwidth]{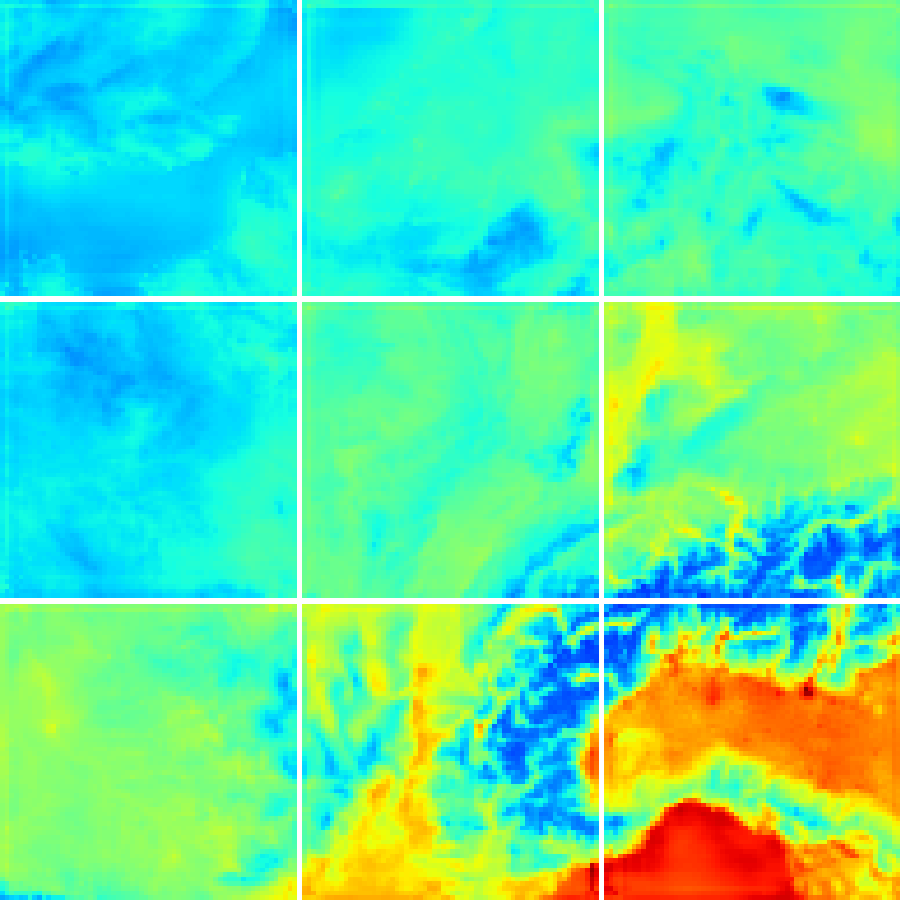}}
    \subfigure[Attack 2) GTBA 5-5]{\includegraphics[width=0.22\textwidth]{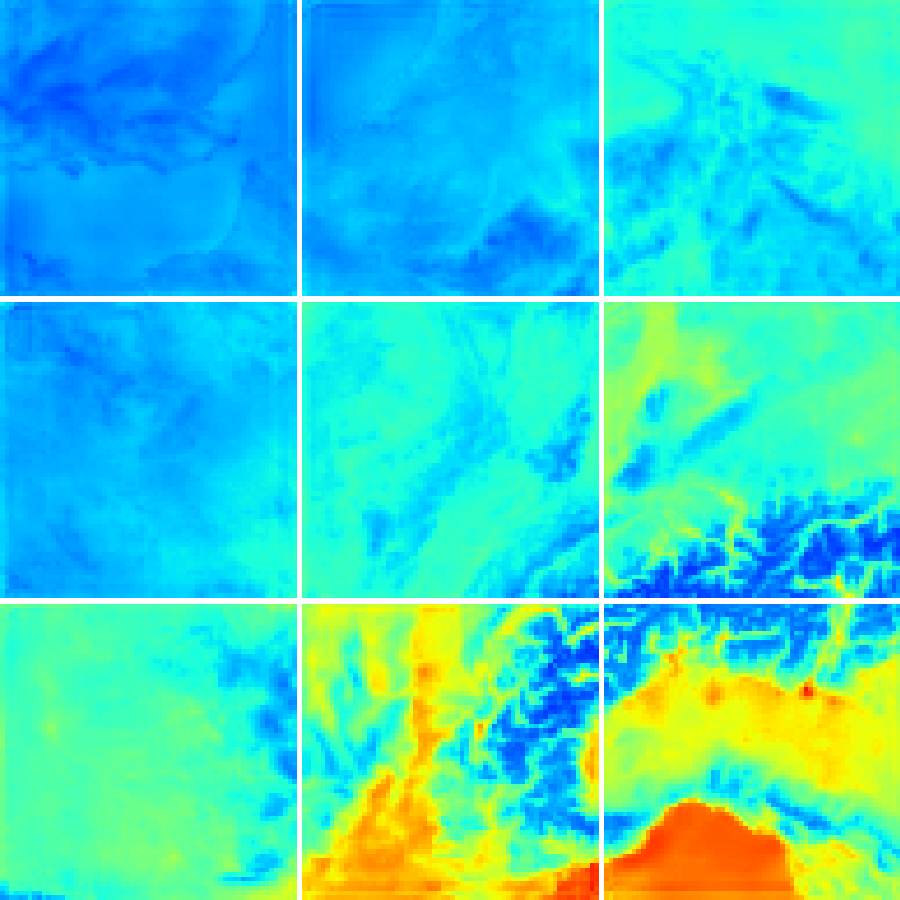}}
    \subfigure[Attack 3) GTBA 3-0]{\includegraphics[width=0.22\textwidth]{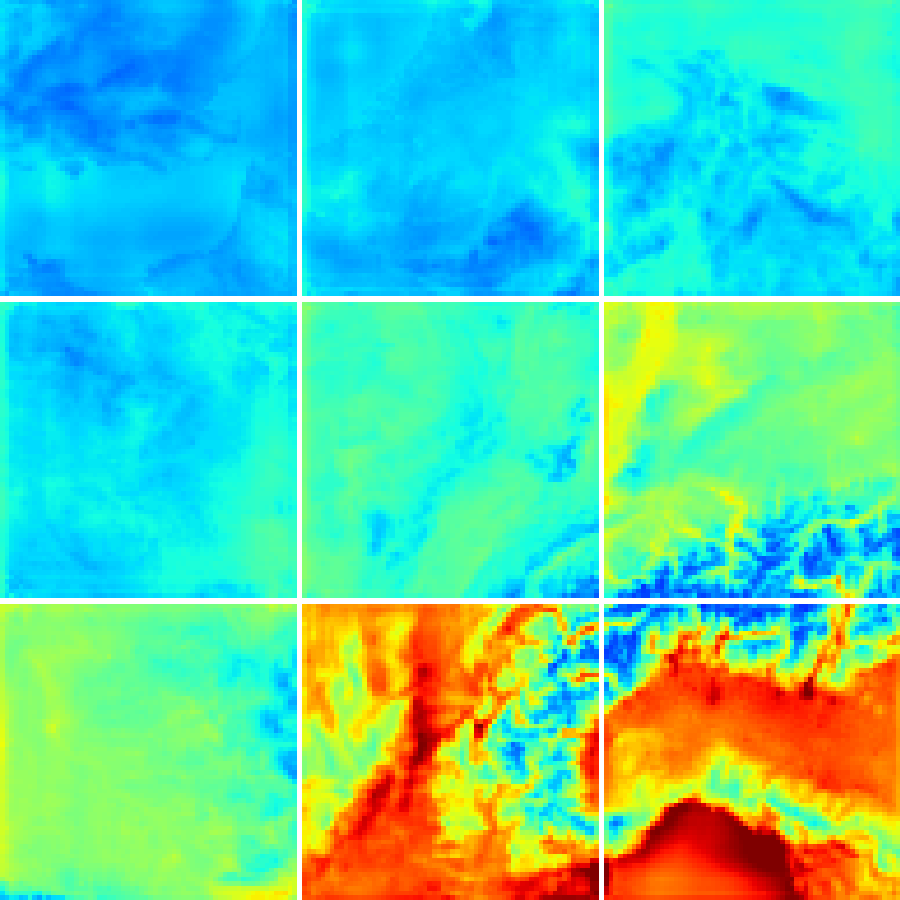}}
    \subfigure[Attack 4) GTBA 3-5]{\includegraphics[width=0.22\textwidth]{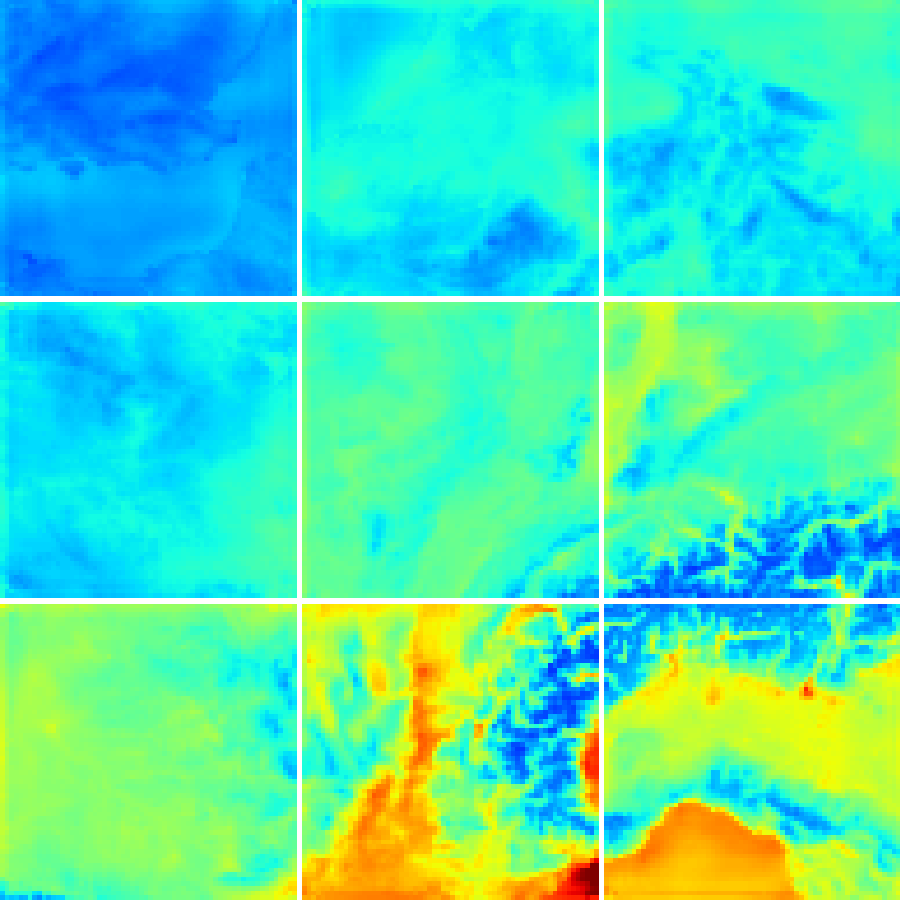}}

    \subfigure[Attack 5) GTBA 1-0]{\includegraphics[width=0.22\textwidth]{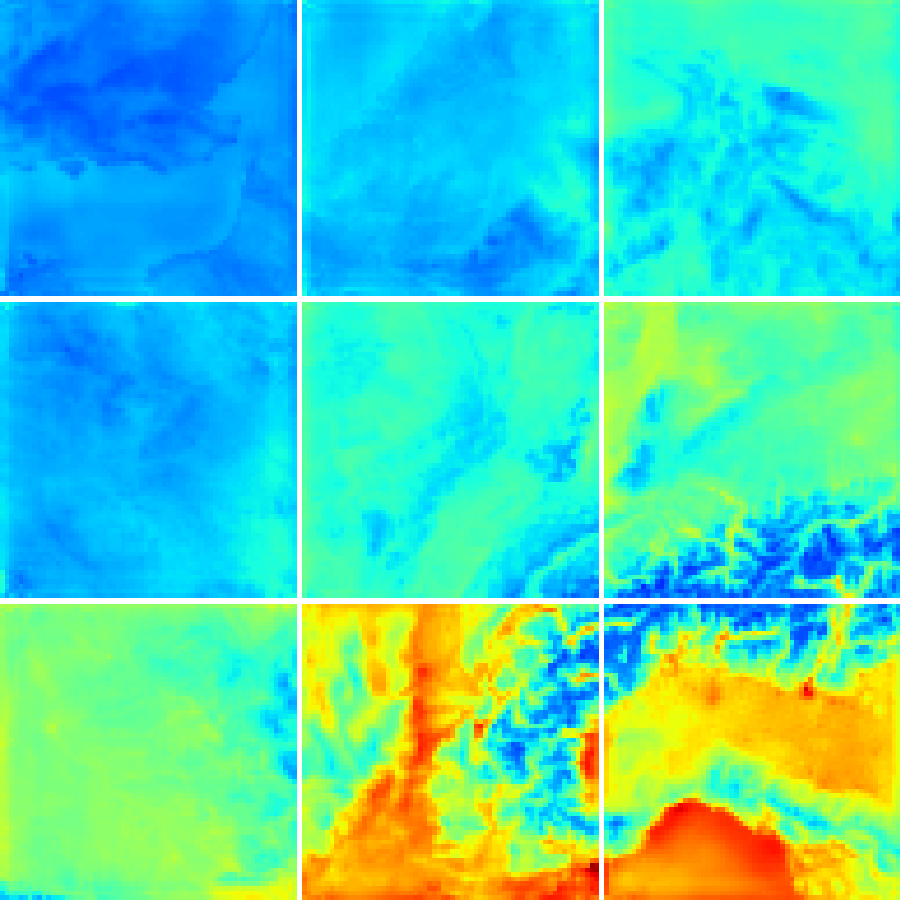}}
    \subfigure[Attack 6) GTBA 1-5]{\includegraphics[width=0.22\textwidth]{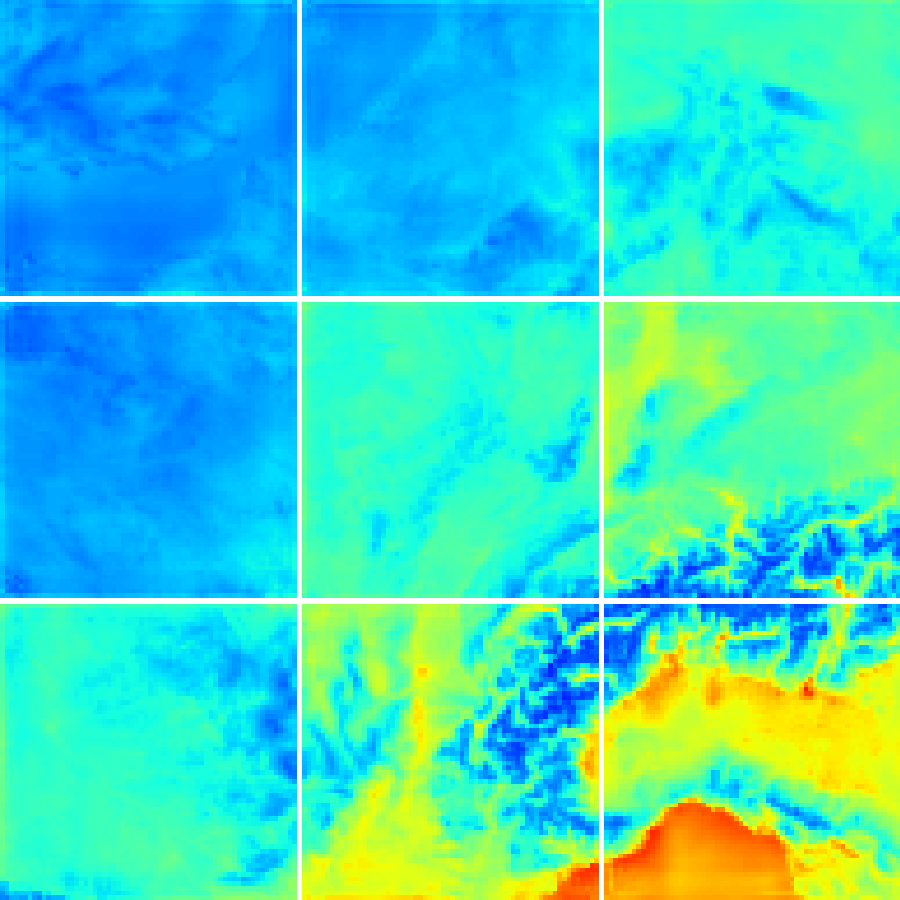}}
    \subfigure[Attack 7) Patch 5-0]{\includegraphics[width=0.22\textwidth]{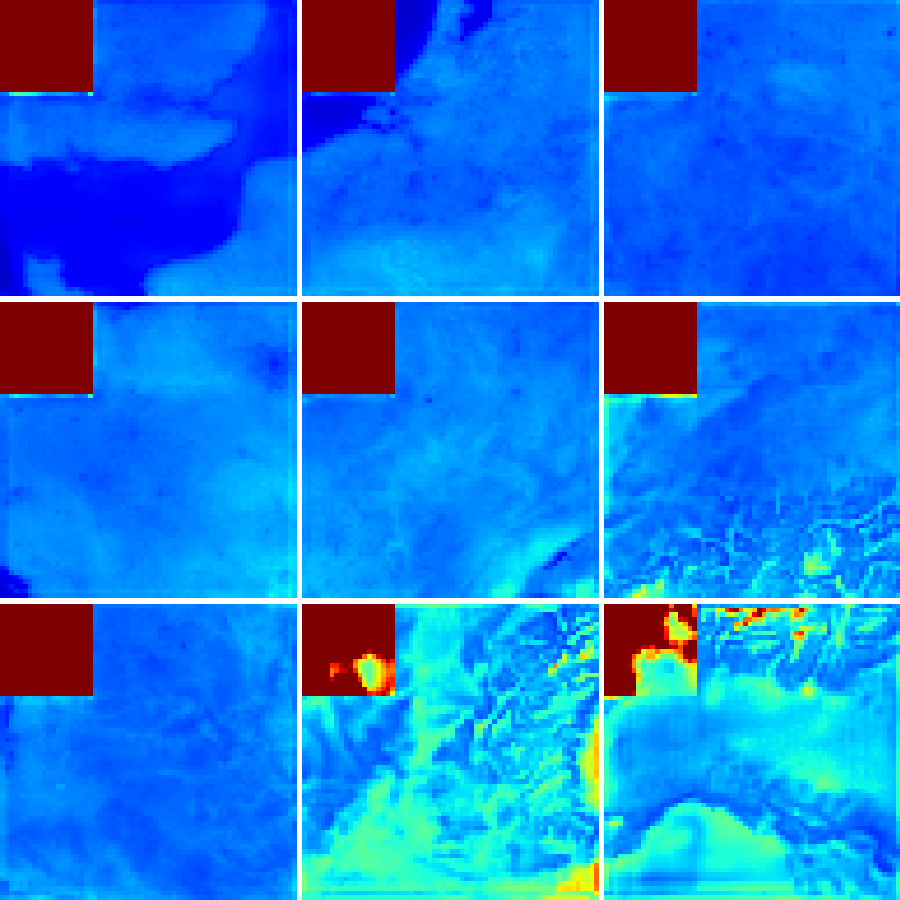}}
    \subfigure[Attack 8) Patch 5-5]{\includegraphics[width=0.22\textwidth]{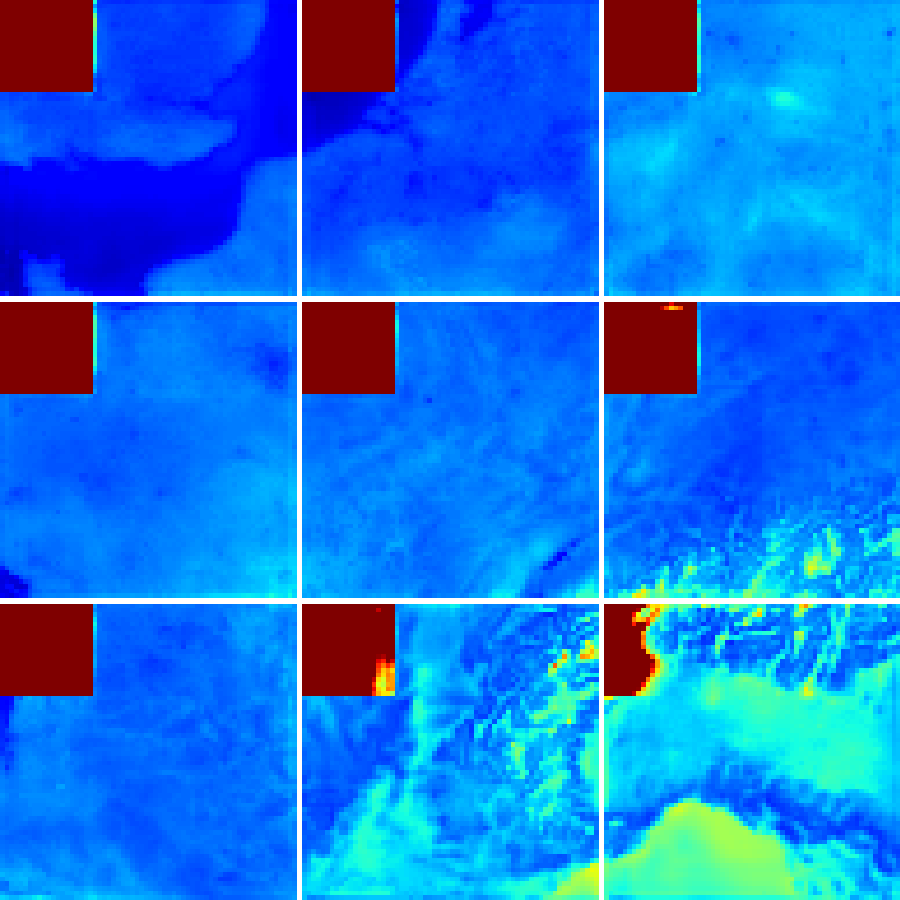}}

    \subfigure[Attack 9) Patch 3-0]{\includegraphics[width=0.22\textwidth]{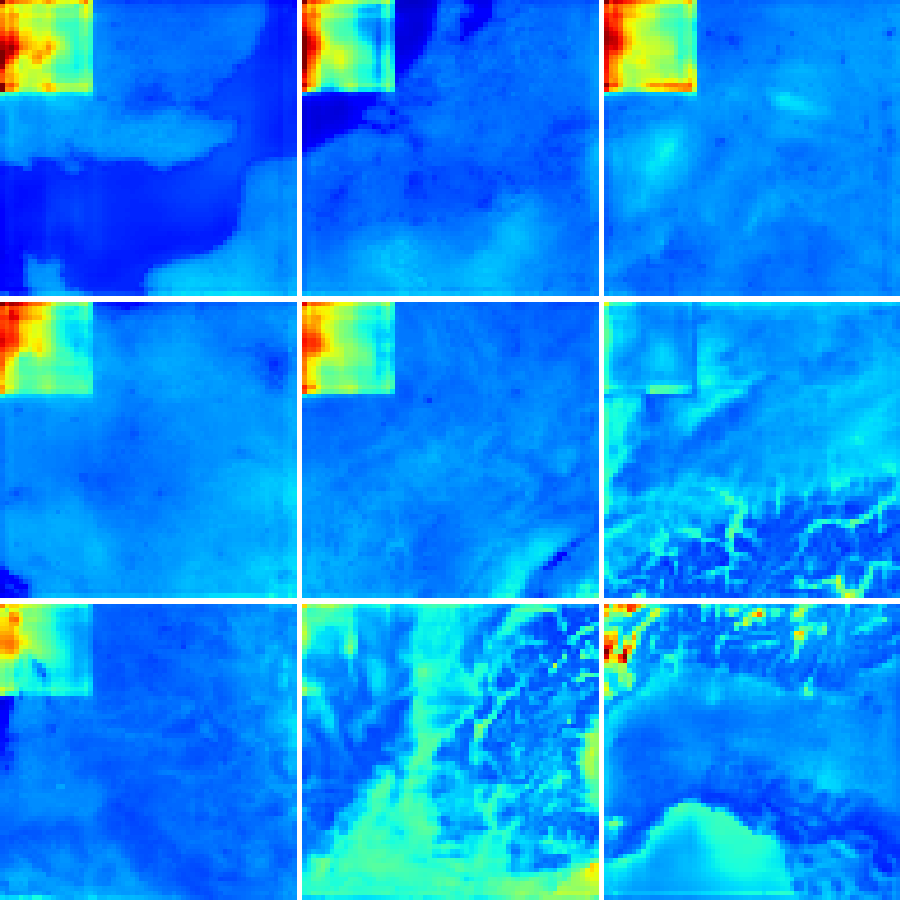}}
    \subfigure[Attack 10) Patch 3-5]{\includegraphics[width=0.22\textwidth]{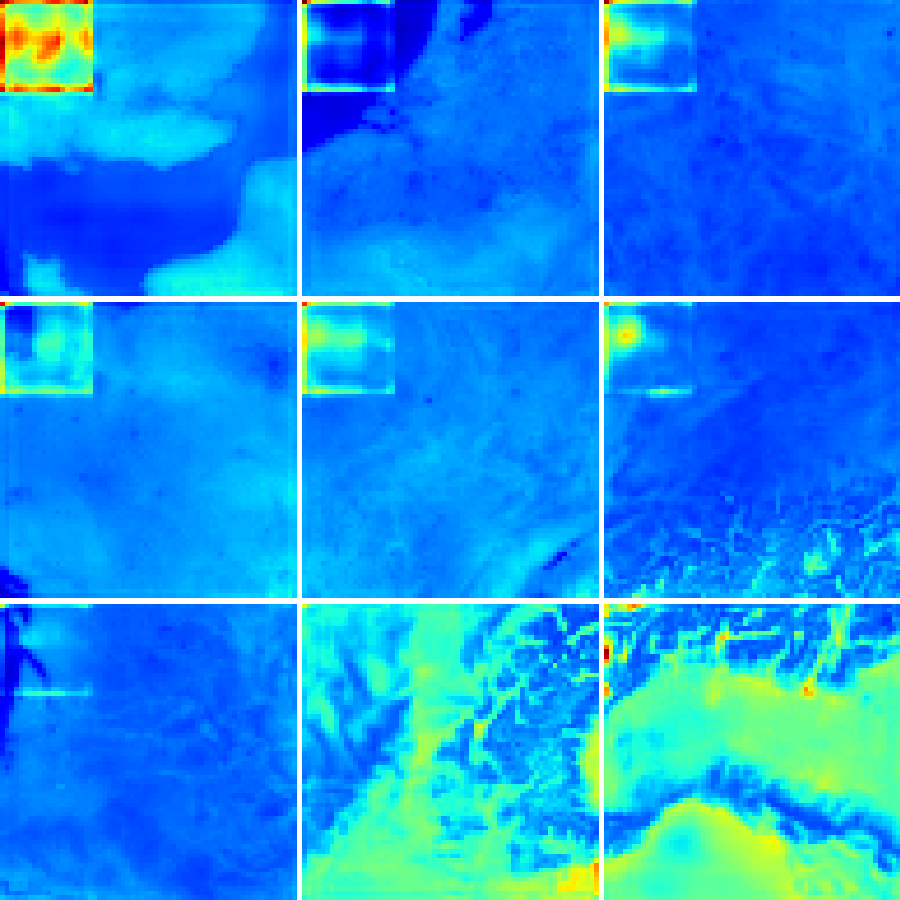}}
    \subfigure[Attack 11) Patch 1-0]{\includegraphics[width=0.22\textwidth]{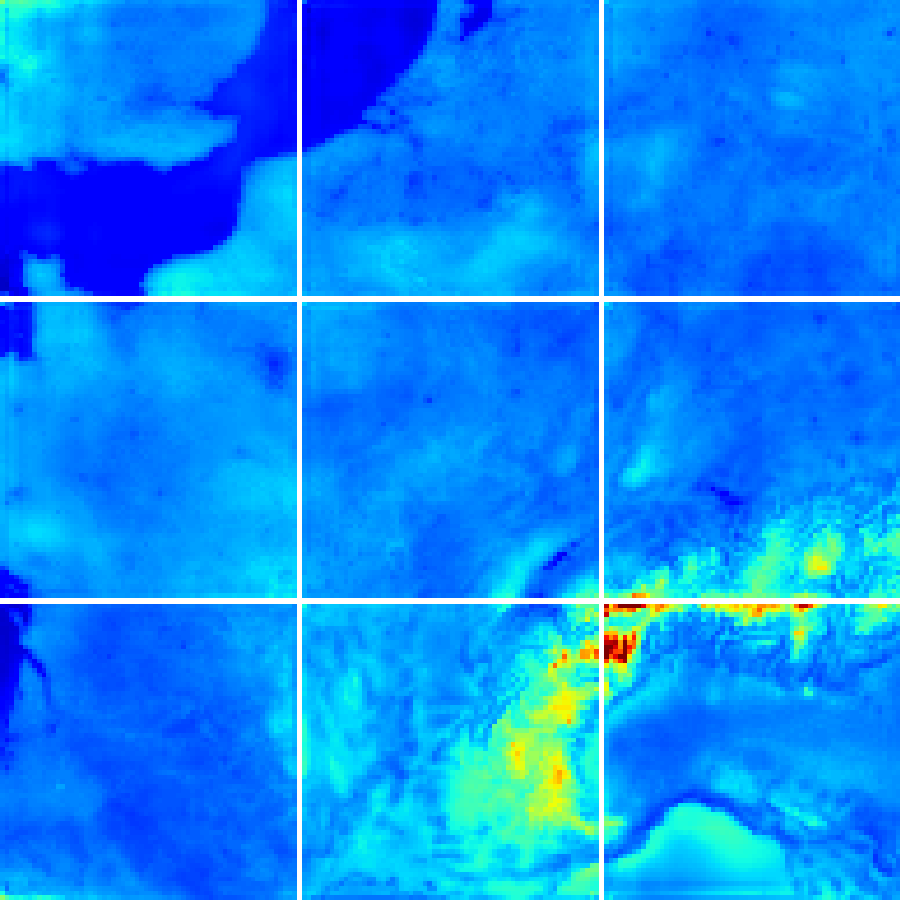}}
    \subfigure[Attack 12) Patch 1-5]{\includegraphics[width=0.22\textwidth]{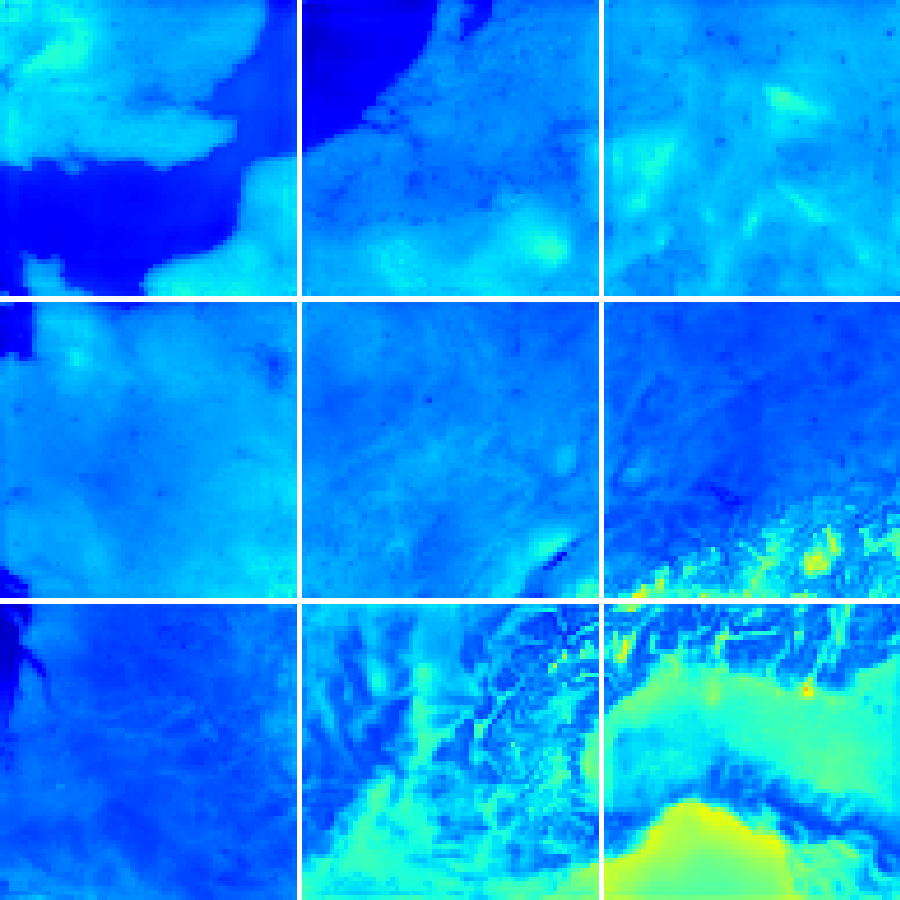}}

    \vspace{0.5em}
    \makebox[\textwidth][c]{%
        \subfigure[No Attacks]{\includegraphics[width=0.22\textwidth]{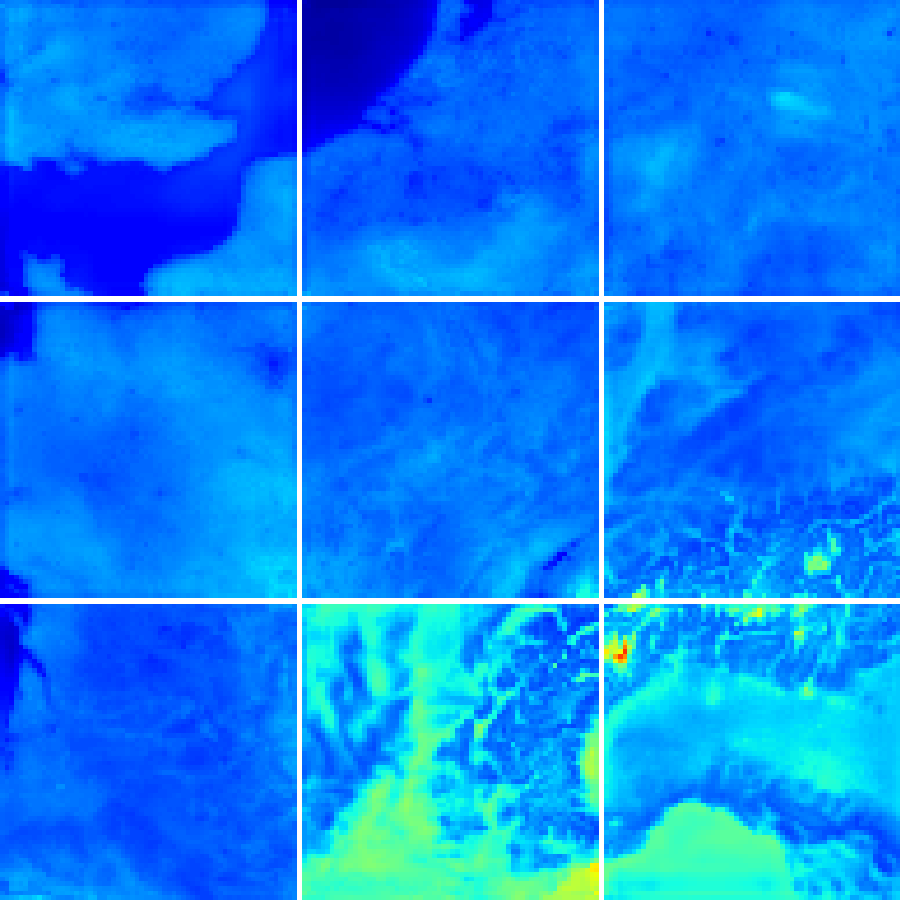}}
    }

    \vspace{0.5em}
    \includegraphics[width=0.4\textwidth]{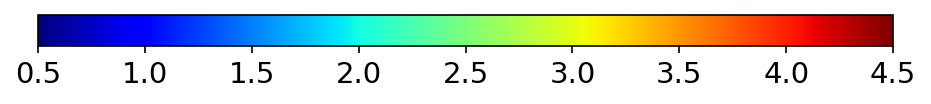}

    {\small \text{Temperature [K]}}

    \caption{
        Comparison of MAE error maps for the legitimate model and all tested attack configurations, evaluated over the complete test set. Each model operates on 9 distinct geographic areas; for each case, a composite image shows the MAE over these regions. The color scale is standardized across all subfigures. Attack names correspond to the configurations detailed in Table~\ref{tab:performance}.
    }

    \label{fig:comparison_mae}
\end{figure*}

This section presents the empirical evaluation of the data poisoning attacks applied to the federated weather forecasting model.

We begin by validating the performance of the clean (non-attacked) federated model. When trained without any  interference, the model demonstrates satisfactory performance across all regional domains, achieving a mean absolute error (MAE) of 1.5127 K and a mean bias of -0.0425 K.

We then examine the effects of adversarial interventions. Two main types of attacks are considered: Global Temperature Bias Attack (GTBA) and Patch Attack. While both target the training labels, they differ in nature and intended effect. GTBA introduces a systematic bias by subtracting the target temperature values by 2 °C, simulating an attempt to conceal warming trends. Patch attacks, on the other hand, introduce localized anomalies by embedding fixed-value patches into the target maps, aimed at studying how localized label corruption propagates through the federated model.

In total, we evaluate 12 attacks, six of each type—across different configurations. Each attack varies in the number of compromised clients (1, 3, or 5) and the injection round (either round 0 or round 5 out of 10 total rounds). All experiments are conducted in a controlled training environment where each round consists of 10 training epochs per client.

Quantitative results are summarized in Table \ref{tab:performance}. We report standard performance metrics—MSE, RMSE, and MAE—as well as SSIM to evaluate perceptual fidelity. To better understand the direction and magnitude of induced biases, we also include mean, minimum, and maximum bias. The clean model serves as a baseline for comparison.

To complement the quantitative analysis, Figure \ref{fig:comparison_mae} visualizes the spatial distribution of the attacks' impact. It shows the mean MAE error maps across the nine regional domains for each attack, highlighting where in the domain space errors are most concentrated.

\subsection{Global Temperature Bias Attack (GTBA)}
GTBA attacks are designed to introduce a systematic shift in the model's learning process by subtracting 2\,K from the ground truth temperature values during training. The impact is evident across all GTBA scenarios, with consistently strong negative mean biases ranging from -1.74\,K to -2.09\,K. This confirms the effectiveness of the attack in shifting the model’s predictions downward.

Several observations emerge from the numerical results:

\begin{enumerate}
    \item Across all tested configurations, the effect on mean bias is substantial and relatively stable, regardless of the number of clients attacked or the injection round. All attacks yield a negative mean bias between -2.0939\,K and -1.7106\,K, closely aligning with the intended 2\,K label reduction. This highlights the severity and reliability of GTBA attacks in introducing large-scale predictive bias.
    
    \item As expected, attacks introduced at earlier rounds lead to more severe degradation. This is evident in Attack 1 (Client 5, Round 0), Attack 3 (Client 3, Round 0), and Attack 5 (Client 1, Round 0), all of which report higher MSE values compared to their corresponding late-round (Round 5) variants.
    
    \item Notably, even a single compromised client is sufficient to induce a significant performance drop. For example, Attack 5 (Client 1, Round 0) yields an MSE of 7.74, close to the highest observed MSE (Attack 3, MSE = 8.60). This indicates that in the case of GTBA, the number of malicious clients is less critical than the timing, and that poisoning even one client can substantially degrade global model quality.
\end{enumerate}

\subsection{Patch Attack}

The Patch Attack introduces a localized, artificial heat anomaly into the label data to degrade forecast quality. This is implemented by adding a patch in the upper-left corner of the target temperature maps for the affected clients. The patch covers approximately 9\% of the spatial domain and is assigned a fixed value equal to the maximum temperature in the training dataset. This setup simulates the injection of a persistent extreme-value signal into a specific region.

\begin{enumerate}
    \item In contrast to GTBA, the number of affected clients plays a critical role in determining the severity of the Patch attack. When 5 clients are compromised, model performance degrades drastically, regardless of whether the attack begins in round 0 or round 5. For instance, Attack 7 (5 clients, round 0) results in an MSE of 22.17, over five times higher than the clean model and more than twice as high as the most damaging GTBA configuration.

    \item When only few clients are attacked, the model shows a high degree of resilience. The resulting errors remain close to the baseline values, with MSE and bias metrics comparable to those of the clean model. This robustness is also visible in the spatial error maps presented in Figure~\ref{fig:comparison_mae}, where the impact of these attacks appears minimal.

    \item A pattern emerges when comparing attack timing across different numbers of compromised clients. With 5 malicious clients (~56\% of participants), early injection (Round 0) produces greater damage than late injection (Round 5), consistent with the intuition that more poisoning rounds lead to greater degradation (Attack 7: MSE = 22.168 vs. Attack 8: MSE = 20.100). However, this relationship reverses when fewer clients are compromised. With 3 malicious clients, Attack 10 (Round 5) exhibits slightly higher MSE (4.756) and lower SSIM (0.725) compared to Attack 9 (Round 0, MSE = 4.742, SSIM = 0.730). A similar inversion occurs with a single malicious client: Attack 12 (Round 5) yields MSE = 4.462 versus Attack 11 (Round 0) with MSE = 4.274.
\end{enumerate}

\subsection{Defense Strategies Against GTBA and Patch Attacks}
To evaluate robustness against adversarial attacks, we tested trimmed mean aggregation with a trim ratio of 0.2. Table \ref{tab:mse_defense_single} reports the comparative performance of FedAvg (FWG) and Trimmed Mean (TM) under clean and adversarial conditions.
Under clean conditions, FedAvg achieves lower MSE (4.124) than TM (4.391), reflecting the expected cost of robust aggregation: trimming discards legitimate updates in benign settings, increasing estimation variance. Against GTBA attacks, TM maintains near-baseline performance across all configurations, with MSE ranging from 4.295 to 4.682 (2–13\% degradation relative to baseline). In contrast, FedAvg degrades substantially under GTBA, with MSE increasing by 88–108\%. This indicates that GTBA produces parameter updates that fall outside the central distribution and are effectively excluded by trimming.
Against Patch attacks, the defensive behavior differs. Attack 7 (5 clients, round 0) overwhelms both aggregation methods, with MSE exceeding 22 for both FedAvg and TM. For Attacks 9 and 11, where fewer clients are compromised, FedAvg remains near baseline (MSE of 4.742 and 4.274 respectively), while TM exhibits degraded performance (MSE of 33.350 and 15.813). This inversion suggests that localized patch perturbations from a minority of clients produce updates that remain within the central parameter distribution, causing trimmed mean to retain adversarial contributions while discarding legitimate ones. The implications of this failure mode are examined in Section \ref{sec:discussion}.

\begin{table}[t]
\centering
\footnotesize
\def\arraystretch{1.15}
\begin{tabular}{lccc}
\toprule
\textbf{Scenario} & \textbf{C–R} & \textbf{MSE (FWG)} & \textbf{MSE (TM)} \\
\midrule
Clean (Legit) & --- & 4.124 & 4.391 \\
\midrule
\multicolumn{4}{l}{\textbf{GTBA}} \\
\quad ↳ Attack 1 & 5–0 & 8.250 & 4.354 \\
\quad ↳ Attack 3 & 3–0 & 8.595 & 4.295 \\
\quad ↳ Attack 5 & 1–0 & 7.742 & 4.682 \\
\midrule
\multicolumn{4}{l}{\textbf{Patch}} \\
\quad ↳ Attack 7  & 5–0 & 22.168 & 22.421 \\
\quad ↳ Attack 9  & 3–0 & 4.742 & 33.350 \\
\quad ↳ Attack 11 & 1–0 & 4.274 & 15.813 \\
\bottomrule
\end{tabular}
\caption{MSE under non-defense (FWG) and defense (Trimmed Mean, TM). C-R represents the couples Client-Round.}
\label{tab:mse_defense_single}
\end{table}

\section{Discussion}
\label{sec:discussion}

The experimental results reveal two fundamentally distinct vulnerability profiles.

GTBA operates as a global, cumulative threat: it requires minimal adversarial presence, scales predictably with injection timing, and produces uniform bias that evades spatial anomaly detection. Its effectiveness stems from exploiting the averaging mechanism itself—a single persistent bias, repeatedly aggregated, eventually dominates the global model. Even a single compromised client can shift predictions by nearly 2 K, demonstrating that the number of malicious participants matters less than their persistence across training rounds. 

Patch attacks, by contrast, function as localized, threshold-dependent perturbations. They require substantial adversarial participation to overcome dilution by honest clients, but when this threshold is crossed, they produce catastrophic and spatially concentrated errors.

A counterintuitive timing pattern emerges in Patch attacks with few malicious clients: early injection causes less damage than late injection. We attribute this to the dilution dynamics of federated averaging. When malicious clients constitute a small fraction of the federation, their influence is progressively suppressed as honest clients push the model toward the true data distribution across subsequent rounds. Early attacks undergo more dilution; late attacks leave insufficient recovery time. This pattern reverses when malicious participants approach 50\% of the federation, as their collective influence rivals that of honest clients, allowing early attacks to embed deeply without effective suppression. As visible in Figure \ref{fig:comparison_mae}, the spatial signature of late-round patches remains clearly apparent, confirming that insufficient post-attack training prevents full recovery.

The evaluation of trimmed mean aggregation as a defensive mechanism reveals a more nuanced and concerning picture. While trimmed mean successfully mitigates GTBA attacks—maintaining near-baseline performance with only 2–13\% degradation compared to FedAvg's 88–108\% increase—its behavior against Patch attacks exposes a fundamental limitation of outlier-based defenses in spatially correlated domains. The key insight is that GTBA produces model updates that manifest as statistical outliers in the aggregation space: the uniform temperature shift creates parameter deviations that fall outside the expected distribution, making them amenable to trimming. Patch attacks, however, exploit the inherent heterogeneity of meteorological data. Because weather patterns naturally vary across geographic regions, the non-IID nature of the data creates a wide distribution of legitimate parameter updates. Localized perturbations—particularly when originating from a minority of clients—generate adversarial contributions that fall within this natural variability rather than outside it.

\section{Conclusions}
\label{sec:conclusions}
This study examines the security risks of federated learning applied to meteorological forecasting, demonstrating that even limited adversarial manipulation can compromise prediction accuracy across large geographic areas. Using the CERRA dataset and a federated U-Net architecture, we evaluate two data poisoning strategies and assess the effectiveness of trimmed mean aggregation as a defense.

The Global Temperature Bias Attack (GTBA) proves particularly effective in this domain: a single compromised client can shift predictions by -1.7 K across the entire network. In operational meteorological systems, such systematic biases could mask warming trends, distort seasonal forecasts, or mislead downstream applications that rely on accurate temperature data, including energy demand planning, agricultural management, and public health warnings. The attack exploits the core aggregation mechanism of federated learning, requiring no sophisticated adversarial capabilities beyond persistent label manipulation.

Patch attacks present a different risk profile, requiring a higher fraction of compromised clients to cause significant damage but producing severe localized anomalies when successful. With five malicious clients, MSE increases by over 400\% and regional biases exceed +3.5 K. This targeted degradation poses concrete operational risks for meteorological applications where localized accuracy is critical, such as frost prediction for agriculture or heat wave alerts for urban areas.

Our evaluation of trimmed mean aggregation reveals that defenses effective against one attack type may fail against another. Trimmed mean successfully mitigates GTBA but amplifies Patch attacks from minority adversaries, as localized perturbations blend with the natural heterogeneity of non-IID weather data. 
This finding is particularly relevant for operational deployment, where defense mechanisms must be both effective and computationally efficient across diverse threat scenarios. The non-IID structure inherent to geographically distributed meteorological observations, while beneficial for model generalization, simultaneously provides cover for adversarial manipulation and complicates the design of robust aggregation strategies.

These vulnerabilities extend beyond meteorology to any federated system with spatially correlated data and geographically distributed participants. Future work should investigate defense mechanisms that leverage physical constraints specific to the forecasting domain, such as spatial continuity or temporal smoothness, and evaluate their computational overhead in realistic operational settings.

\section*{Declarations}

\subsection*{Funding}
This research was partially funded and supported by the following Projects:
\begin{itemize}
\item  Project SERICS
(PE00000014) under the MUR National Recovery and Resilience Plan funded by the European Union - NextGenerationEU
\item European Cordis Project ``Optimal High Resolution Earth System Models for Exploring Future Climate Changes'' (OptimESM),
Grant agreement ID: 101081193
\item ISCRA Project ``AI for weather analysis and forecast'' (AIWAF)
\end{itemize}

\subsection*{Authors' contributions}
KC designed the federated learning models and their attacks, implemented the experiments, and analyzed the results. FM provided the climate datasets, designed the U-Net architecture and offered domain expertise.

\subsection*{Statements}
The authors declare no competing interests.

\subsection*{Code and data availability}
The code relative to the presented work is archived at the following GitHub repository \hyperlink{https://github.com/TryKatChup/Adv-attacks-FL-meteorology}{https://github.com/TryKatChup/Adv-attacks-FL-meteorology}

The data used in this study were obtained from the Copernicus archive hosted on the Climate Data Store, openly accessible at \hyperlink{https://cds.climate.copernicus.eu/datasets/}{https://cds.climate.copernicus.eu/datasets/} 

\bibliography{bibliography}

\end{document}